\documentclass[conference]{IEEEtran}
\IEEEoverridecommandlockouts
\usepackage{cite}
\usepackage{amsmath,amssymb,amsfonts}
\usepackage{algorithmic}
\usepackage{graphicx}
\usepackage{textcomp}
\usepackage{xcolor}
\usepackage{url}
\usepackage[utf8]{inputenc}
\def\BibTeX{{\rm B\kern-.05em{\sc i\kern-.025em b}\kern-.08em
    T\kern-.1667em\lower.7ex\hbox{E}\kern-.125emX}}
\begin{document}

\title{Deep Learning Based Motion Planning For Autonomous Vehicle Using Spatiotemporal LSTM Network\\
}

\author{\IEEEauthorblockN{1\textsuperscript{st} Zhengwei Bai}
\IEEEauthorblockA{\textit{School of Electronic and Information Engineering} \\
\textit{Beijing Jiaotong University}\\
Beijing, China \\
zwbai@bjtu.edu.cn}
\and
\IEEEauthorblockN{2\textsuperscript{nd} Baigen Cai, 3\textsuperscript{rd} Wei ShangGuan*, 4\textsuperscript{th} Linguo Chai}
\IEEEauthorblockA{\textit{School of Electronic and Information Engineering} \\
\textit{State Key Laboratory of Rail Traffic Control and Safety}\\
\textit{ BERC of EMC and GNSS Technology for RT} \\
\textit{Beijing Jiaotong University}\\
Beijing, China \\
*Corresponding author, wshg@bjtu.edu.cn}
}

\maketitle

\begin{abstract}
Motion Planning, as a fundamental technology of automatic navigation for autonomous vehicle, is still an open challenging issue in real-life traffic situation and is mostly applied by the model-based approaches. However, due to the complexity of the traffic situations and the uncertainty of the edge cases, it is hard to devise a general motion planning system for autonomous vehicle. In this paper, we proposed a motion planning model based on deep learning (named as spatiotemporal LSTM network), which is able to generate a real-time reflection based on spatiotemporal information extraction. To be specific, the model based on spatiotemporal LSTM network has three main structure. Firstly, the Convolutional Long-short Term Memory (Conv-LSTM) is used to extract hidden features through sequential image data. Then, the 3D Convolutional Neural Network(3D-CNN) is applied to extract the spatiotemporal information from the multi-frame feature information. Finally, the fully connected neural networks are used to construct a control model for autonomous vehicle steering angle. The experiments demonstrated that the proposed method can generate a robust and accurate visual motion planning results for autonomous vehicle.
\end{abstract}


\section{Introduction}
Motion planning is a fundamental technology for autonomous driving vehicles. Although numerous motion planning methods have been proposed to handle variety specific application environment\cite{b1,b2,b3}, the dynamic and complex traffic situation makes state-of-art general motion planning methods not work well.  To make the motion planning strategy more general and robust is the primary task of perception based autonomous vehicle control method.

Over the past years,the apply of Deep Neural Network to estimate the perception-action methods in real-time control architecture has recently been demonstrated to have a dramatic advanced capacity in scalability and performance of autonomous vehicle technology(AVT)\cite{b4,b5}. Moreover, deep learning methods greatly enhances the adaptability of the automatic driving model to complex environments and is not limited to model-based control, which means that the generalization of the control architecture is greatly enhanced\cite{b6}.

In 2016 Nvidia proposes the DAVE2 end to end training framework, based primarily on the camera's picture as input, to output driving commands by training the CNN-FNN structure\cite{b7}. Although the conventional end to end training method can implement a simple automatic driving model construction simply and quickly, this kind of simple deep CNN-FNN training method not only has a huge model volume demand, but also has a black box control approach that lacks theoretical support.

With the advance and development of Conv-LSTM, the Conv-LSTM can not only establish timing relationships like LSTM, but also can characterize local spatial features like CNN\cite{b8}. In 2017, Baidu adopted an automated driving training model based on the Conv-LSTM Network structure\cite{b9}. By using the picture feature extraction capabilities from convolutional network and the memory ability from LSTM networks, the proposed network model can learn the motion planning more accurately.

It is well known that 3D Convolutional Neural Network(3D-CNN) has outstanding capabilities for the extraction of spatial features\cite{b10}. Thus, compared with using Conv-LSTM alone, Conv-LSTM combined with 3D-CNN (named as Spatiotemporal LSTM Network) will be more conducive to extracting the spatiotemporal information from the environment. However, this structure is rarely used in motion planning area so far.

In this paper, we use the Spatiotemporal LSTM Network to train the motion planning model for autonomous vehicle. This paper is organized as follows. Section 2 described our network architecture, Section 3 performed the experiments and analysis, and conclusions are given in the last section.
\section{Approach}
In this section, we explain the core ideology of Conv-LSTM and describe the architecture of our proposed Spatiotemporal LSTM Network which combines Conv-LSTM and 3D Convolutional Neural Network for motion planning.

\begin{figure*}[htbp]
\centerline{\includegraphics[width=18cm]{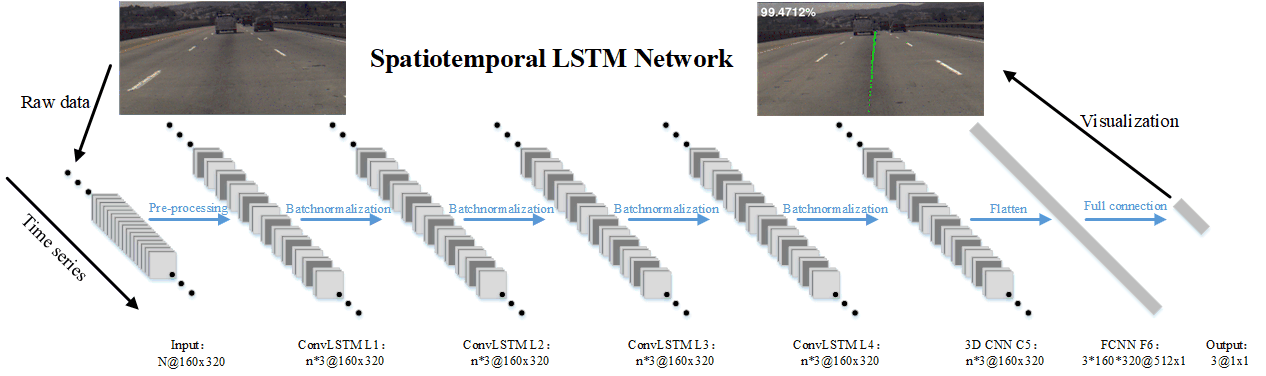}}
\caption{The architecture of the Spatiotemporal LSTM Network}
\label{fig1}
\end{figure*}

\subsection{Proposed Network Architecture}
The overall network architecture of the proposed approach for autonomous vehicle motion planning using deep learning is shown in Fig. 1. For the whole network processing, we input the multi-frame picture segment into the system and finally get the steering output after passing the spatiotemporal LSTM network.Since the original information we have collected is the single-frame picture information collected by the camera, we need to perform preprocessing before entering the system.In the pre-processing process, single-frame picture information is combined in time series to form a multi-frame picture segment based on time series.After that, the multi-frame image segment based on time series is input into the system, which is followed by 4 layers of Conv-LSTM layers and a 3D-CNN layers. Finally the output of the motion planning result works out through 2 layers of FCNN.

\subsection{Long-Short Term Memory(LSTM)}
In the traditional Recurrent Neural Network(RNN), the Vanishing Gradient Problem(VGP) is a major obstacle for RNN. The emergence of LSTM solved the VGP for RNN and greatly promoted the development of RNN. The basic block of LSTM is shown as Fig. 2.

\begin{figure}[htbp]
\centerline{\includegraphics[width=7.5cm]{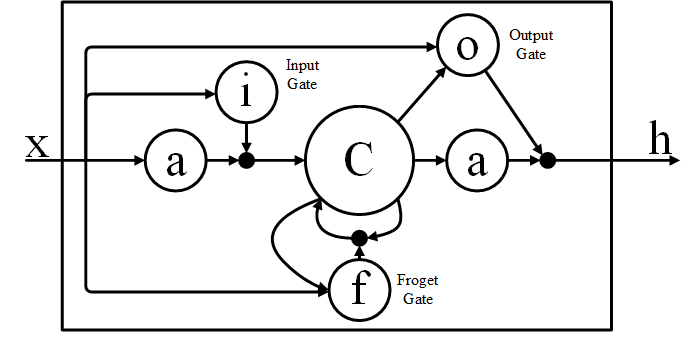}}
\caption{The architecture of LSTM block. Firstly, given an input, the cell data is generated by Input Gate(IG), Forget Gate(FG) and the output of last block.Then the cell data is fed into FG, OG and the activation function of next block. Finally, the output of this block is generated by OG and activation function of the cell.}
\label{fig2}
\end{figure}

LSTM (Long Short-Term Memory) is a long-term and short-term memory network. It is a time-recursive neural network and is suitable for processing and predicting important events of relatively long intervals and delays in time series.Three gate were placed inside a cell, called the input gate, the forget gate, and the output gate. A message enters the LSTM network and can be used to determine if it is useful. Only the information that meets the algorithm's certification will remain, and the inconsistent information will be forgotten through the forget Gate.The autopilot learning behavior itself is also a learning model based on time series and selective memory. Thus ,the LSTM is suitable to train a autonomous driving model.

\subsection{Data Preprocessing}

In the process of constructing the deep learning network, it is very important to analyze and preprocess the original data.The raw data used in this paper is the picture frame output by the camera on the car.In order to make these time-series-based picture frames better adapted to the proposed network to achieve better training performance, we need to preprocess them which is shown as Fig. 3. In addition, grouping single-frame images based on time can also allow input data to contain richer temporal and spatial data.

\begin{figure}[htbp]
\centerline{\includegraphics[width=7.5cm]{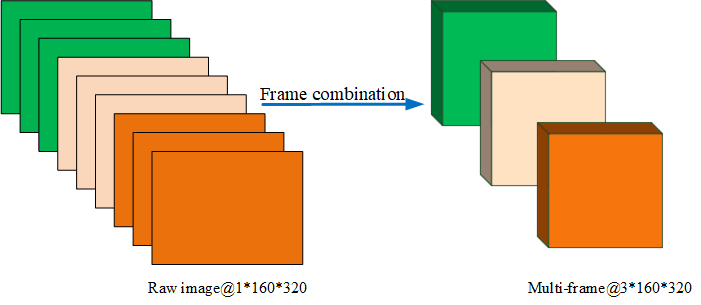}}
\caption{The data pre-processing approach which is used to transform the single frame image to multi-frame data.}
\label{fig3}
\end{figure}

First, through the frame combination processing of the input image, the discrete image frame  is input to the convolutional LSTM as time-serial image segment for processing. Compared to traditional, we use shorter time series to generate multi-frame picture fragments. This is because taking into account the time-related efficiency is a very important factor in the movement of vehicles, so here we use a three-frame combination to generate timing clips and feed it into our proposed spatiotemporal network.

\subsection{Spatiotemporal feature extraction}

Our proposed spatiotemporal LSTM network is based on convolutional LSTM (Conv-LSTM), 3D convolutional neural network(3D-CNN) and fully connected neural network(FCNN) to extract the spatiotemporal features of time-serial image segment. In our proposed network, 4 Conv-LSTM layers are used and each of them are combined with a batch normalization layer(BN)\cite{batchnorm} to prevent the problem of low training efficiency caused by data distribution offset in deep networks.
For traditional LSTM, it is often used in sequence-based processing systems because it has sequence-based data analysis and processing capabilities.However, the traditional LSTM does not have a good ability to learn continuous pictures. Fortunately, the proposal of Conv-LSTM makes the temporal memory training based on multi-frame picture segments greatly enhanced.The output of the previous layer is the input of the next layer. The difference is that after the convolution operation is added, not only the timing relationship can be obtained, but also features can be extracted like a convolutional layer to extract spatial features. In this way, the spatiotemporal characteristics can be obtained.the core process of Conv-LSTM is shown as Fig. 4

\begin{figure}[htbp]
\centerline{\includegraphics[width=7.5cm]{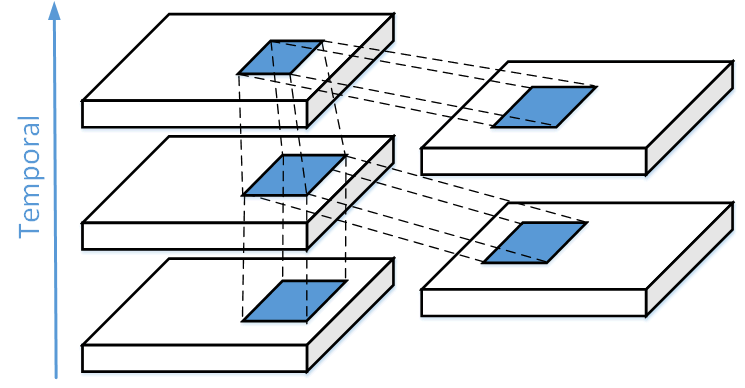}}
\caption{The core architecture of  Conv-LSTM.}
\label{fig4}
\end{figure}
As can be seen from the figure, the connection between the input and each gate at this time is replaced by convolution by the feedforward type, and the convolution operation is also replaced between the state and the state.
In terms of the autonomous driving process, it is not a discrete behavior like object recognition but a continuous response through the recognition of the environment. So, in our proposed method, we use a four layers Conv-LSTM structure to extract the temporal hidden feature information and help the model learn better from the data.The working principle of Conv-LSTM used in this paper can be shown as follows:
\begin{equation}
i_{t} = \sigma (W_{xi} * X_{t} + W_{hi} * H_{t-1} + W_{ci} \circ C_{t-1} + b_{i})
\end{equation}
\begin{equation}
f_{t} = \sigma (W_{xf} * X_{t} + W_{hf} * H_{t-1} + W_{cf} \circ C_{t-1} + b_{f})
\end{equation}
\begin{equation}
C_{t} = f_{t} + i_{t} \circ tanh(W_{xc} * X_{t} + W_{hc} * H_{t-1} + b_{c})
\end{equation}
\begin{equation}
o_{t} = \sigma (W_{xo} * X_{t} + W_{ho} * H_{t-1} + W_{co} \circ C_{t} + b_{o})
\end{equation}
\begin{equation}
H_{t} = o_{t} \circ tanh(C_{t})
\end{equation}
Where $i_{t}$, $f_{t}$, $c_{t}$ represent the value from input gate, forget gate and output gate at moment $t$ respectively, at the same time, the $o$ means the $Hadamard$ product between two matrix. The entire system of equations represents the state generating mechanism from moment $t-1$ to moment $t$.
It should be noted that the $X$, $C$, $H$, $i$, $f$ and $o$ are all three-dimensional tensors. The latter two dimensions represent the spatial information of the rows and columns.(The first dimension should be the time dimension). So it can predict the characteristics of the center grid based on the characteristics of the surrounding points in the grid.
Considering that in the deep network training process, the low-level network updates the parameters during training and causes the distribution of upper-level input data to change.So, at the input of each layer of the network, a batch normalization layer is inserted, that means, a normalization process is performed first, and then the next layer of the network is entered. The main process of batch normalization is shown as follows:
\begin{equation}
\mu_{B} \gets \frac{1}{m} \sum_{i=1}^{m} x_{i}
\end{equation}
\begin{equation}
\sigma_{B}^{2} \gets \frac{1}{m} \sum_{i=1}^{m} (x_{i} - \mu_{B})^{2}
\end{equation}
\begin{equation}
\widehat{x_{i}} \gets \frac{x_{i} - \mu_{B}}{ \sqrt{\sigma_{B}^{2} + \epsilon }}
\end{equation}
\begin{equation}
y_{i} \gets \gamma\widehat{x_{i}}  + \beta \equiv BN_{\gamma,\beta}(x_{i})
\end{equation}
Where $x_{i}$ represents the element of the input batch, $m$ represents the size of the batch, so $\mu$ represents the overall mean of the batch. $B$ is the variance of the batch, which represents the stability of this batch of data. $\widehat{x_{i}}$ is the result of batch standardization of batch data. Finally, the parameters $\gamma$ and $\beta$ allow $x_{i}$ to be transformed and reconstructed to obtain the results of normalized data without changing the feature distribution.
For the image process methods, he traditional way of using 2D-CNN to operate on time-serial images is generally to use CNN to identify each frame of the video. This method does not take into account the inter-frame motion information in the time dimension.Using 3D-CNN can better capture the temporal and spatial characteristics of video. Traditional 2D-CNN and the 3D-CNN convolution operation on the images are shown as Fig. 5,6:
\begin{figure}[htbp]
\centerline{\includegraphics[width=7.5cm]{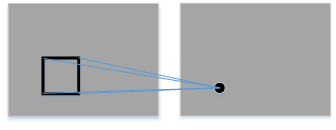}}
\caption{2DCNN convolution operation on the image using 2D convolution kernel.}
\label{fig5}
\end{figure}

\begin{figure}[htbp]
\centerline{\includegraphics[width=7.5cm]{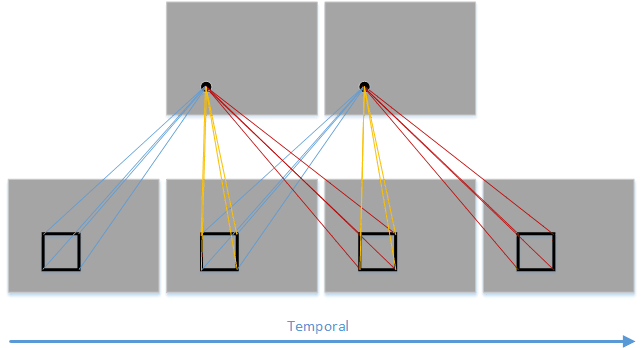}}
\caption{3D-CNN convolution operation on the image using 3D convolution kernel.}
\label{fig6}
\end{figure}

For 3D-CNN, the convolution operation has a time dimension of 3, that means a convolution operation on three consecutive frames of images.The above 3D convolution is to build a cube by stacking multiple consecutive frames, and then use the 3D convolution kernel in the cube.
In this structure, each feature map in the convolutional layer is connected to multiple adjacent consecutive frames in the previous layer, thus capturing motion information.The value of a position of a convolutional map is obtained by convolving local receptive fields at the same position of three consecutive frames one level above.
\subsection{Generation for motion planning}
In our proposed method, for the last step, we use a fully connected neural network to predict the motion planning value of the autonomous vehicle. For the network structure, first a dense layer with 512 neurons is added behind the output layer of last BN layer. Meanwhile, the 0.5 dropout rate is added and using the LeakyReLU activation function shown as follow:

\begin{equation}
LeakyReLU(x) =\left\{
\begin{array}{rcl}
-0.2 * x, &{x      <      0}\\
x, & {x \leq 0 }
\end{array} \right.
\end{equation}

We can see from the function, compared with conventional activation function ReLU which sets all negative values to zero, in contrast, Leaky ReLU assigns a non-zero slope to all negative values so that the overall data can be more completely passed to the next layer of the network. After that, a dense layer with just one neuron is used to merge the all information to the output as the motion planning value. The loss function used in model is shown as follow:

\begin{equation}
MSE(x, \hat{x}) = \frac{1}{n_{samples}} \sum_{i=0}^{n_{samples}-1} (x_{i} - \hat{x_{i}})^{2}
\end{equation}

Mean square error(MSE) is often used as a loss function because it can be derived. The result of the calculation is the sum of the squares of the difference between the predicted value and the true value of the network. Ultimately, the output value of a fully-connected network is the expected motion planning value we expect. In addition, we have also visualized the results of motion planning so that we can see the quality of our network more intuitively.

\section{Experiments and Analysis}
\subsection{Experimental platform and database}
In order to evaluate the proposed method, we used the Comma-dataset\cite{comma}, which consist of over 80GB raw image data and vehicle actual state data which are collected from the vehicle-based sensors respectively. For the model training platform,the experimental environment is equipped with Inter(R) Core(TM) i7-6700 CPU @ 3.4GHz, 32 GB RAM, and a NVIDIA GTX 980 GPU. For the software the experiment was build by Keras\cite{keras} with Python.
The images with three channels, 320*160 pixels captured by the camera installed in the front windshield of test vehicle in the frequency of 20Hz. The comma-dataset also collected the steering angle, angular velocity and linear velocity through the vehicle sensors with the units deg, deg/min and km/h respectively.The raw image and steering angle data are used into our propose method.

\subsection{Model training}
In the selection of experimental parameters,taking into account the application scenario,we choose 3 frames per fragment for the input frame dimension to ensure the real-time performance of motion result.For the Conv-LSTM layer, we set the kernel size to 3 and the filter first to 64 but finally 8, because we find that a lower filter will help the 3D-CNN perform better.For the 3D-CNN layer we set the kernel size to 3 and the filters to 3.
Then we use a flatten layer to transform the 3-D output to a 1-D input for the FCNN. In terms of the fully connected layers we use the LeakyReLU activation function which is shown as Fig and the a dropout with the rate 0.5 which is used to prevent the overfitting. Besides, we use the MSE and Adam as our loss function and optimizer respectively. During the training, we use the batch size of 20\cite{batch} to get a faster converge speed.

\subsection{Results and Analysis}
The visualization experimental result proposed in this paper is shown as  Fig. 7. The blue curves represent the real driving motion. At the same time, the green curves represents the planning motion generated by our method. Obviously, the green and blue curves are very well-fitted which means that our proposed method works well for autonomous vehicle.
\begin{figure}[htbp]
\centerline{\includegraphics[width=8.5cm]{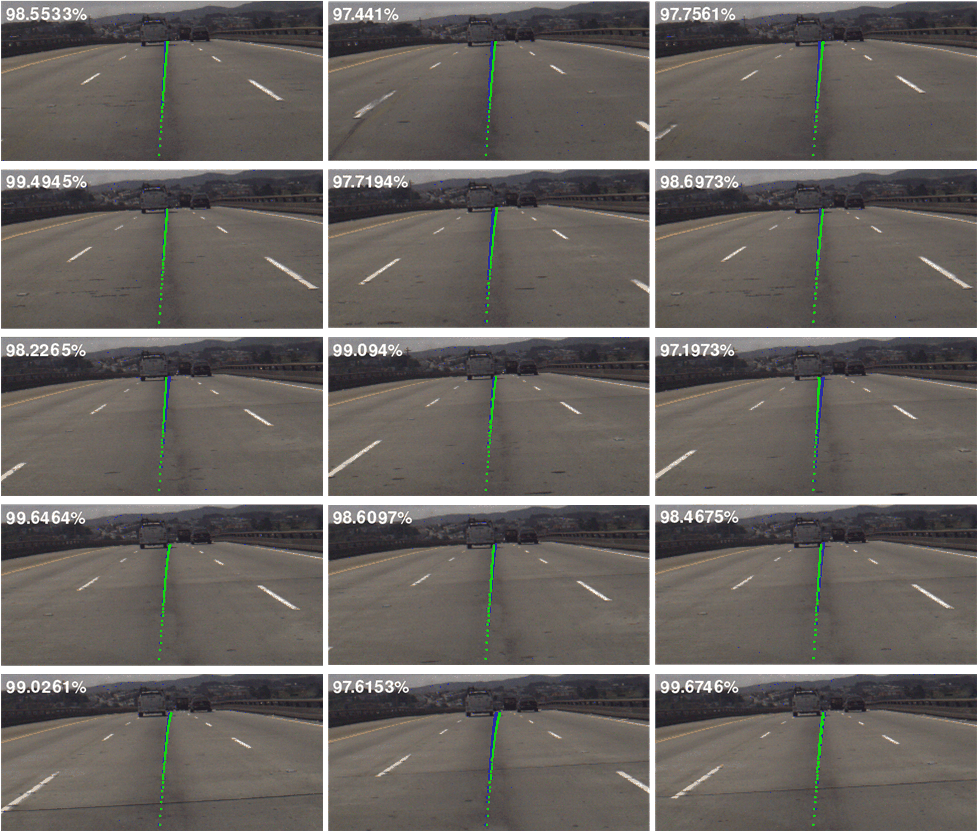}}
\caption{The visualization result of propose motion planning model experiments}
\label{fig7}
\end{figure}
During the training process we find that in each epoch of the training (generally from the 5th start), the loss value first declines rapidly, then rebounds a bit and then steadily decreases until the epoch ends. After analysis, we think this is because the deep spatiotemporal neural network will always start training from the state of lower loss of the previous epoch during training. However, our proposed training data is time-based sequence data, which means maybe the model generate overfitting data based on the antecedent data from the previous moment first and then, through the optimizer, the model will keep finding the real general minimize state again.
In order to further evaluation our method, we evaluate both our and Hotz's method\cite{comma}. We constructed a neural network based on 2D-CNN and FCNN according to the Hotz's scheme.Then we randomly selected a section of data on the highway and verified our and Hotz's models respectively. Finally, the evaluation results is shown in Fig. 8.
\begin{figure}[htbp]
\centerline{\includegraphics[width=8.5cm]{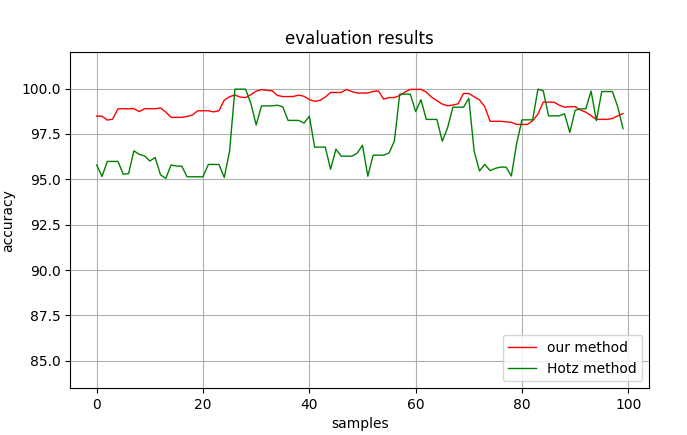}}
\caption{Experiments curve of our and Hotz's methods}
\label{fig8}
\end{figure}
Through the figure we can see that both our and Hotz's method can make great motion planning for the real-time traffic environment. But it is clear that Hotz's approach has greater stability fluctuations. In comparison, our method not only has a better average evaluation performance but also has a  lower average variation  performance. So our method can be more reliable and stable while generating the autonomous motion planning results.

\section{Conclusion}
This paper presents a spatiotemporal LSTM network for autonomous vehicle motion planning which is able to generate the steering motion reaction by exploiting the spatiotemporal information. The convolutional LSTM has the ability to learn the time-serial features about the traffic environment, Meanwhile, the 3D-CNN can extract the spatial information. Experimental results verified the validity and stability of our proposed deep learning method. In the future we plan to implement the spatiotemporal network in more complex autonomous vehicle control scenario which will prompt us to understand the autonomous driving mechanism in depth.

\section*{Acknowledgment}
This paper is supported by National Key Research and Development Program of China (2016YFB1200100), Beijing Natural Science Foundation (4172049), National Natural Science Foundation of China Monumental Projects (61490705, 61773049).

\end{document}